\documentclass{article}

\PassOptionsToPackage{numbers, square}{natbib}

\usepackage[final]{neurips_2021}

\usepackage[utf8]{inputenc} % allow utf-8 input
\usepackage[T1]{fontenc}    % use 8-bit T1 fonts
\usepackage{hyperref}       % hyperlinks
\usepackage{url}            % simple URL typesetting
\usepackage{booktabs}       % professional-quality tables
\usepackage{amsfonts}       % blackboard math symbols
\usepackage{nicefrac}       % compact symbols for 1/2, etc.
\usepackage{microtype}      % microtypography
\usepackage{xcolor}         % colors
\usepackage{framed}
\usepackage{amsmath}
\usepackage[capitalise,noabbrev,nameinlink]{cleveref}

\usepackage{lipsum}
\usepackage{listings}
\usepackage{graphicx}
\graphicspath{ {./images/} }

\title{An Interactive Visualization Tool for Understanding Active Learning}

\author{%
  Zihan Wang 
  \qquad Jialin Lu 
  \qquad Oliver Snow 
  \qquad Martin Ester \\
  School of Computer Science\\
  Simon Fraser University\\
  Burnaby, BC, Canada \\
}

\begin{document}

\maketitle

\begin{abstract}
Despite recent progress in artificial intelligence and machine learning, many state-of-the-art methods suffer from a lack of explainability and transparency. The ability to interpret the predictions made by machine learning models and accurately evaluate these models is crucially important. In this paper, we present an interactive visualization tool to elucidate the training process of active learning. This tool enables one to select a sample of interesting data points, view how their prediction values change at different querying stages, and thus better understand when and how active learning works. Additionally, users can utilize this tool to compare different active learning strategies simultaneously and inspect why some strategies outperform others in certain contexts. With some preliminary experiments, we demonstrate that our visualization panel has a great potential to be used in various active learning experiments and help users evaluate their models appropriately. 
\end{abstract}

% keywords can be removed
% \keywords{Interactive Human Centered Machine Learning\and Active Learning \and Visualization}

\section{Introduction}

Recent advances in machine learning (ML) and artificial intelligence (AI) have been mainly focused on making models more accurate and efficient. However, there has been an increasing focus on improving the explainability and interpretability of models as well as considering the human users of these models \citep{schmidt2020interactive}. It is crucially important for humans to be able to explain the predictions made by ML models especially in some sensitive domains, such as automated financial investing, autonomous driving, and healthcare management. There are still many ML applications where state-of-the-art models can achieve strong predictive power, but the gain in accuracy comes at the cost of transparency, and the decision reached lacks interpretability \citep{mononet}. Meanwhile, several algorithms may achieve highly similar performance based on some experimental setting, which makes it difficult to evaluate their effectiveness and distinguish how they truly work. 

Active Learning (AL) is an approach that has proven useful when labeled training data is scarce and unlabeled data is plentiful, as is the case in many real-world settings \citep{multicriteria}. AL works by selecting unlabeled examples to be labeled by an oracle (usually a human annotator), in order to train an accurate model with the least labeled examples \citep{Settles}. The majority of theoretical work in AL has taken place in classification problems, with many fundamental and successful algorithms developed \citep{uncertainty, Abe1998QueryLS}. To evaluate AL, the most common approach is to plot \textit{accuracy} by number of queries, where we expect accuracy to improve as more samples are queried. A few recent studies also adapt some similar algorithms for regression \citep{NEURIPS2018_dc4c44f6, WU201990}, and the usual performance measure is \textit{Mean Squared Error} (MSE). However, the information reflected from either the accuracy or MSE plots is both limited and potentially biased to users. For example, many previous papers presented highly similar accuracy curves \citep{zhan2021multiplecriteria,beck2021effective,ash2020deep}, yet a number of contradictory findings have been recorded in past work \citep{beck2021effective}. Hence, users may not be able to well-distinguish different behaviors between diverse AL algorithms singly from the accuracy plot.

% Although AL often involves human interaction during the querying phase, unfortunately there have not been much published work on explaining how AL works through interactive visualization. In addition, the information reflected from either the Accuracy or the MSE plots is both limited and potentially biased to users. Even if two AL algorithms present highly similar curves in the gain of accuracy, we should be cautious to conclude that they are equally effective. In fact, many previous papers presented highly similar accuracy curves \cite{zhan2021multiplecriteria,beck2021effective,ash2020deep}, yet a number of contradictory findings have been recorded in past work, including large differences in accuracy of certain baselines and contradictory conclusions on the relative performance of certain algorithms \cite{beck2021effective}. To sum up, users may not well-distinguish different behaviors between diverse AL algorithms singly from the accuracy plot.

In this paper, we propose a novel interactive visualization tool for people to better understand why and how AL works on certain classification and regression tasks. By using dimension reduction techniques to create a 2-D feature embedding for visualization, users are able to select a proportion of interesting test data points and view how their prediction values change according to more queries. The prediction values for selected points are arranged to a 2-D mesh-grid plot called \textbf{prediction-change} plots with each pixel showing the prediction differences. Compared to normal accuracy curves, our new tool illustrates how accuracy changes for a specific subgroup of data. We demonstrate that with our approach, users could compare different AL strategies simultaneously, get an intuition on why certain algorithms do not work on some test points, and evaluate them appropriately. 

%In addition, we demonstrate that this tool is adaptive to most AL classification and regression experiments, as it only requires the test dataset and prediction values as input. 

\section{Related Work}

% \subsection{Active Learning and Interactive Visualization}
\paragraph{Active Learning and Interactive Visualization}
To the best of our knowledge, there is no established Human-Centered or interactive visualization tool for understanding how and why AL works. The closest we can get is an online prototype "Active Learner" \citep{cloudera}. It serves as an exhibition and only gives settled results for three different datasets and four AL strategies. Other notable examples include work by \citet{inbook}, in which they introduced a user interface for querying samples to be labeled, specifically for image recognition problems. In addition, \citet{Iwata2013ActiveLF} proposed an AL framework for interactive visualization which selects objects for the user to re-locate so that they can obtain the desired visualization to their preference. Crucially, these studies did not investigate explainability or visualization for the evaluation of AL methods.

% \subsection{Evaluation of Active Learning}
\paragraph{Evaluation of Active Learning}
Many successful AL querying methods have been developed and shown to outperform some baseline (often random sampling). Comparing accuracy/MSE over querying more samples is the most common approach for evaluation. Other metrics are sometimes used as well such as AUC and F1-score \citep{Ramirez}. However, focusing on accuracy could lead to unexpected and unwarranted conclusions. For example, \citet{Ramirez} claimed that AL often improves accuracy at the expense of recall, which means that the observed improvement in accuracy is not fully due to \textit{effective} learning. %A seeming advantage of AL over random sampling based on accuracy can be misleading.
To draw a proper evaluation of AL, accuracy plots are usually not sufficient and other facets should be considered, such as AL experimental setting, labeling efficiency, and redundancy \citep{beck2021effective}. Our proposed visualization tool can be helpful in many of them. 

\section{Prediction-change Plots}

\subsection{Definition of the prediction-change Check}
We consider a general AL process, in which an oracle provides a fixed number of labels. The initial labeled dataset is $D_{l}=\{(x_{1},y_{1}),...(x_{M},y_{M})\}$ and the unlabeled data pool is $D_{u}=\{x_{1},...,x_{N}\}$ where each instance $x_{i} \in \mathbb{R}^{d}$ is a $d$-dimensional feature vector. $y_{i}$ is the ground truth value of $x_{i}$. In each iteration, the model chooses $D'=\{(x',y')\}$ as the queried new sample. 

Let $f$ be the mapping function of our model (either a classifier or a regressor) and $\theta$ be its parameters (if any). For a test sample $x_{\text{test}}$ from the test set $D_{\text{test}} = \{(x_{\text{test}},y_{\text{test}})\}$, the prediction of the model trained on the original training set $D_{l}$ is $\tilde{y}_{\text{test}} = f( x_{\text{test}} ;\theta_{D_{l}})$. The prediction of our model trained with new added samples is $\tilde{y}^{'}_{\text{test}} = f( x_{\text{test}} ;\theta_{D_{l} + D'})$. Single $\tilde{y}_{\text{test}}$ may or may not change.

\textbf{The first prediction-change check}:
For some of the test samples, we expect the new prediction value of $\theta_{D_{l} + D'}$ is different from the old one of $\theta_{D_{l}}$:
\begin{equation}
| f( x_{\text{test}} ;\theta_{D_{l} + D'}) - f( x_{\text{test}} ;\theta_{D_{l}})| \neq 0
\label{eq:check1}
\end{equation}
If there is no change for all test samples, this may suggest our new queried sample does not provide new information for the model.

\textbf{The second prediction-change check}:
For some of the test samples $x_{\text{test}}$ we selected, we expect the prediction of $\theta_{D_{l} + D'}$ is closer to the ground truth $y_{\text{test}}$, than to the prediction of $\theta_{D_{l}}$:
\begin{equation}
| f( x_{\text{test}} ;\theta_{D_{l} + D'}) - y_{\text{test}}| < | f( x_{\text{test}} ;\theta_{D_{l}}) - y_{\text{test}} |
\label{eq:check2}
\end{equation}
The prediction values can possibly be worse for some test samples. Overall, the sum of all prediction differences from our selected test samples compared to all ground truth values is expected to be smaller, which means the model is consistently performing better.

\subsection{Design of the Interactive Visualization Panel}

Regardless of how the AL experiment is conducted, this interactive visualization tool only requires two inputs: 1. the test dataset $D_{\text{test}}$; 2. the prediction values on this set over a number of AL queries (could be stored in a matrix). We let users dynamically select "similar" and interesting test samples to look at how their prediction values change.

Firstly, Principal Component Analysis (PCA) is conducted and the first two PCs are extracted. Then, we plot these PCs to create a 2-D feature embedding for visualization and depict a general structure of all test samples, if there exists some similarity between the test samples. Finally, users can select a targeted region of points that they want to inspect. The interactive panel will automatically generate the \textit{prediction-change} plots based on a number of (default is 20) nearest points. In detail, prediction values for selected points are arranged to a 2-D mesh-grid plot: \textbf{x-axis} represents the querying process (unit could be single or batch queries); \textbf{y-axis} represents the indices of selected test samples; \textbf{each pixel} shows prediction differences according to some criteria. Derived from definitions in $3.1$, we propose three types of prediction-change plots based on the change of the prediction values between:

\textbf{1. current model vs. original model} (\cref{eq:check1}): If the model is progressively improving and learning from the queried instances, we expect the prediction values to be gradually more and more different. This can be reflected by more colorful mesh-grid graphs.

\textbf{2. current model vs. previous model} (\cref{eq:check1}): For some strategies, we expect later queried samples to bring less effect on changing the model. This can be reflected by lighter colors on the plot.

\textbf{3. current model vs. ground truth} (\cref{eq:check2}): As querying more samples, we expect the model to make better predictions that are closer to ground truth. This can also be reflected by a lighter plot.

Notice that for classification, pixels in the prediction-change plot have finite number of colors. For example, prediction-change values can only be $\{1,0,-1 \}$ for a binary classification with labels $\{ 0, 1\}$. In regression, these values are continuous, which suggests pixel colors to be gradient-based.

\section{Experiments}

% \subsection{Setting}
\paragraph{Settings}

\begin{figure}[h]
\centering
 \includegraphics[width=0.94\linewidth]{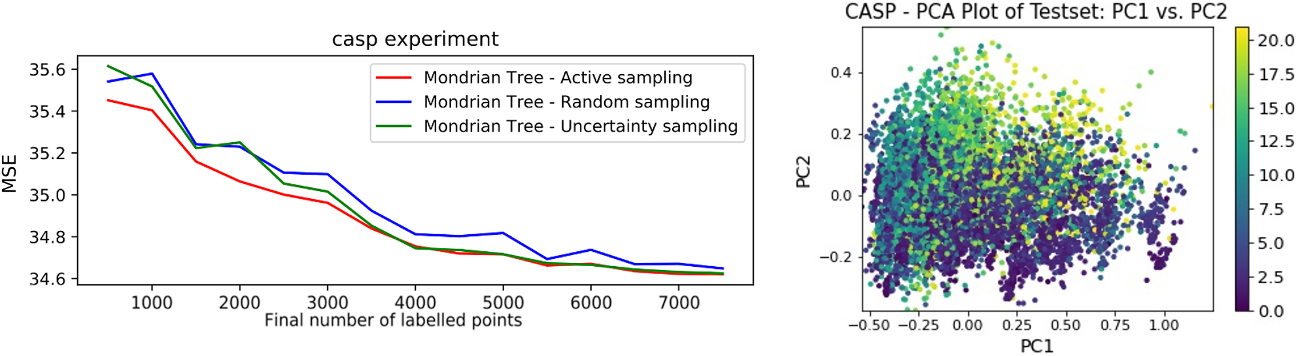}
 {
 }
 \caption{\textbf{Left}: MSE plot of the AL experiment. 
          \textbf{Right}: PC1 vs. PC2 of all test samples}
 \centering
 \label{fig1}
\end{figure}

Because gradient colors are more recognizable in regression, we conducted our experiments on a regression problem with real world data: CASP \citep{CASP}, which contains 45,730 instances in total. We randomly selected 9,730 (\%21) of them as the test set. For the AL process, we started with an empty training set, batch queried 500 at each iteration, and stopped after 15 batches. Fig. \ref{fig1} (left) shows the MSE plot, and Fig. \ref{fig1} (right) illustrates the PCA plot for all 9,730 test samples.

The regressor comes from \citet{NEURIPS2018_dc4c44f6}, which used purely random trees, specifically, Mondrian Trees. We compared the performance of three basic querying strategies: its tree-based active algorithm (al), a naive uncertainty sampling (uc) version of its active algorithm, and random sampling (rn).

% \subsection{Results of the Interactive Visualization Panel}
\paragraph{Results}

\begin{figure}[h]
\centering
 \includegraphics[width=0.876\linewidth]{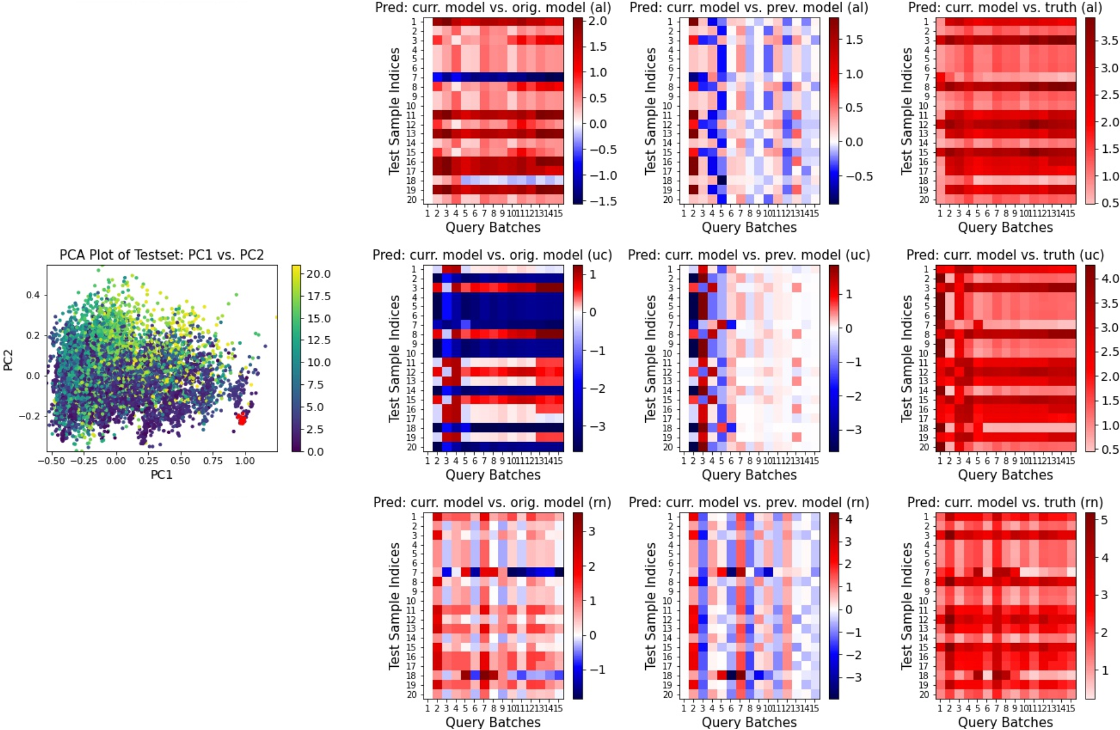}
 {
 }
 \caption{Prediction-change plots for three AL strategies on examples with small values}
 \centering
\label{fig2}
\end{figure}

There are some obvious clustering patterns in the PCA plot. To start off, we select the points at the bottom right corner with the smallest values. The red points in the PCA plot of Fig. \ref{fig2} mark the points we selected. Also, $3*3$ panel plots are provided for the prediction-change plots of three strategies ("al", "uc", and "rn") in each row. A complete user-interface is in Appendix Fig. \ref{fig3}. 

Looking at the \textit{first} prediction-change plots across three querying algorithms, it is fairly clear that they perform differently in changing the regressor model, even though they have similar MSE curves in Fig. \ref{fig1}. "al" performs more regularly than "uc" or "rn", with consistently larger values (18 out of 20 purely red rows) or smaller values (1 out of 20 purely blue rows) compared to the initial model. However, unlike this diverse performance on the low-value group, three algorithms show almost uniform performances for the large value group (plots shown in Appendix A.2 Fig. \ref{fig4}). To investigate potential reasons, we further check the batched queried samples and their true values (data shown in Appendix A.3 Fig. \ref{fig5}). Compared to the original CASP data distribution, all three algorithms show similar data distributions for the first 500 batch-queried samples. Although "al" and "uc" seem to query more large value points (near 20), unfortunately this does not help the Mondrian Tree model perform significantly better in making predictions. One inference we can draw is that the model does not learn well from the queried samples for fitting large value points or data points from a sparse area. 

In the \textit{second} prediction-change plot of "uc", we see dark pixels at first and quite light, nearly white color pixels after 5 query batches. This suggests that "uc" tends to change the model dramatically at the first stage and has less effect in later queries. The \textit{third} prediction-change plots consistently showed red pixels, which intuitively makes sense because the dark points we selected have small truth values near 0. The tree regressor tends to fit the structure with larger predictions than the truth. 

%"uc" seems to perform worst after initial queries. Those initially selected samples might badly contribute to training the tree regressor, as "uc" queries the most informative samples.

% The dark points we selected have small truth values and the tree regressor tends to fit the structure with larger predictions than the truth. This pattern is consistently reflected by red pixels in the \textit{third} prediction-change plots. "uc" seems to perform worst after initial queries. Those initially selected samples might badly contribute to training the tree regressor, as "uc" queries the most informative samples. In later batch queries, "uc" changes the model less, which is reflected from nearly white color pixels in its \textit{second} prediction-change plot approximately after 5 query batches. We gain an intuition that "uc" tends to bring dramatic changes to the model in the beginning and bring less effect later. By contrast, "al" and "rn" seem to fluctuate irregularly at each stage.

%To further investigate how each algorithm works, we could look into the data properties of these points and test on other data groups (e.g., mixed large and small values). 

\section{Conclusion}

Human-Centered AI and ML are receiving increased attention from researchers, although there have not been many established Human-Centered techniques for AL. In this paper, we have proposed an interactive visualization tool for AL based on prediction values, inspired by the background that there is limited explainability potential from traditional accuracy/MSE plots. We introduce a simple definition of prediction-change checks, which is applicable in both classification and regression settings. Users are able to select a sample of data points and get a better intuition on how the predictions of those points change through AL. Our initial experiments on real-world regression tasks indicate several interesting insights and suggest that there are many more that this tool could reveal. 

\paragraph{Limitations and Future Work}
It is easier for users to play around with this visualization panel and draw conclusions if there are clear clusters shown on the PCA plot. If test samples do not show any clear association or cluster groups, the AL task will be more complicated because of no clear decision boundaries, which further makes our prediction-change plots less distinguishable. There are many additional questions we can ask by running these panel plots in different experimental settings. Future work involves testing on different data sets, which leads to many interesting avenues for further exploration - e.g., do certain AL methods work well in certain data subgroups? %Another direction is to conduct experiments on classification problems with developed querying strategies. 

%%%%%%%%%%%%%%%%%%%%%%%%%%%%%%%%%%%%%%%%%%%%%%%%%%%%%%%%%%%%

\bibliography{references} 
\bibliographystyle{unsrtnat}

\appendix

\section{Appendix}

\subsection{User Interface of the Prediction-change Plots}

This is a sample user interface of the visualization panel, using the example from Fig. \ref{fig2}. 

\begin{figure}[h]

 \includegraphics[width=0.99\linewidth]{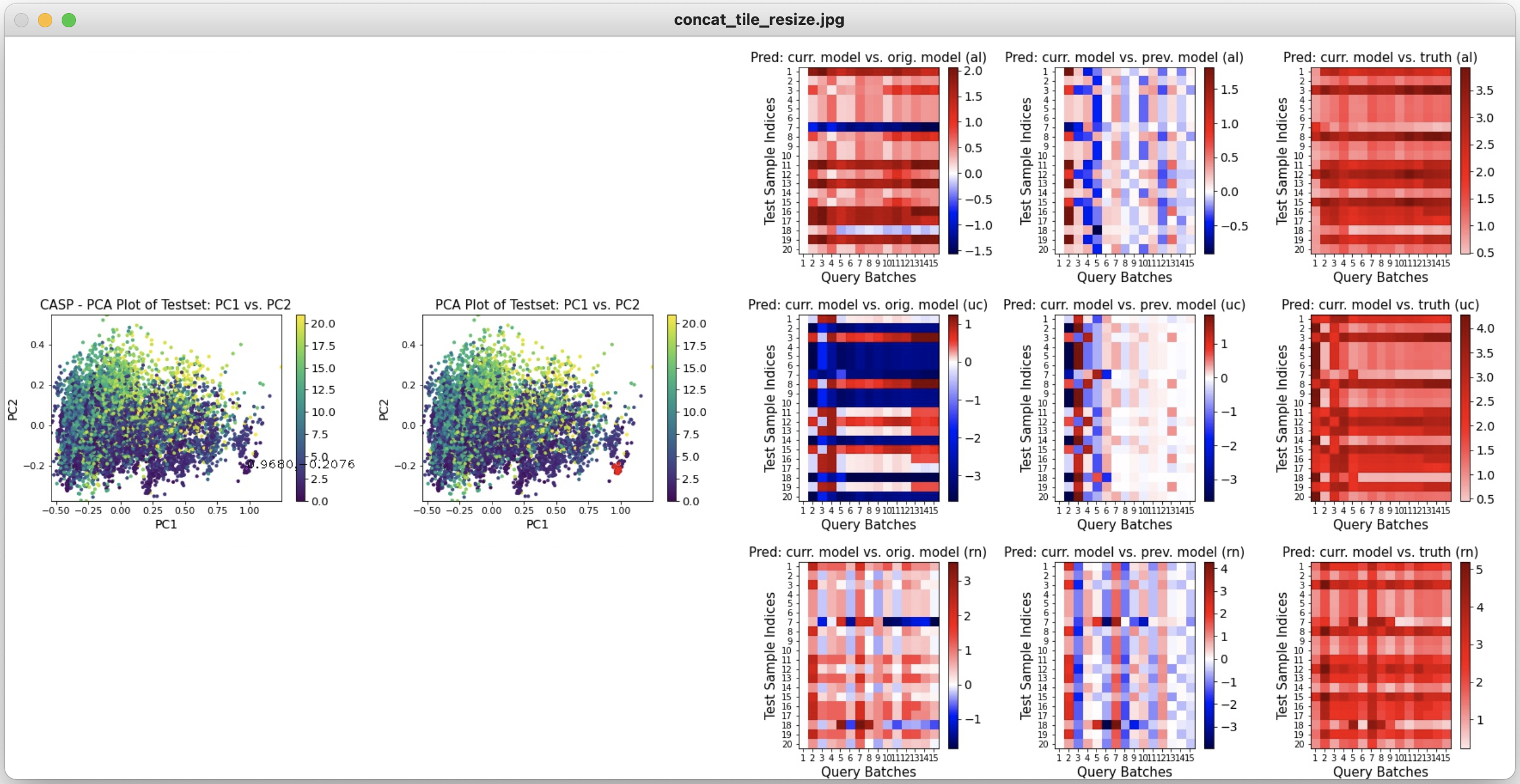}
 {
 }
 \caption{The actual user interface of Fig. \ref{fig2}}
 \centering
\label{fig3}
\end{figure}

\subsection{Prediction-change Plots for Large Value Group}

In this example, the red points we selected contain large values (light yellow color in PCA plot; closed to 20). Three querying alrogithms ("al", "uc", "rn") show highly similar performance for all prediction-change plots. The tree model might struggle to regress this large value group, regradless of what new samples being queried. 

\begin{figure}[h]

 \includegraphics[width=0.99\linewidth]{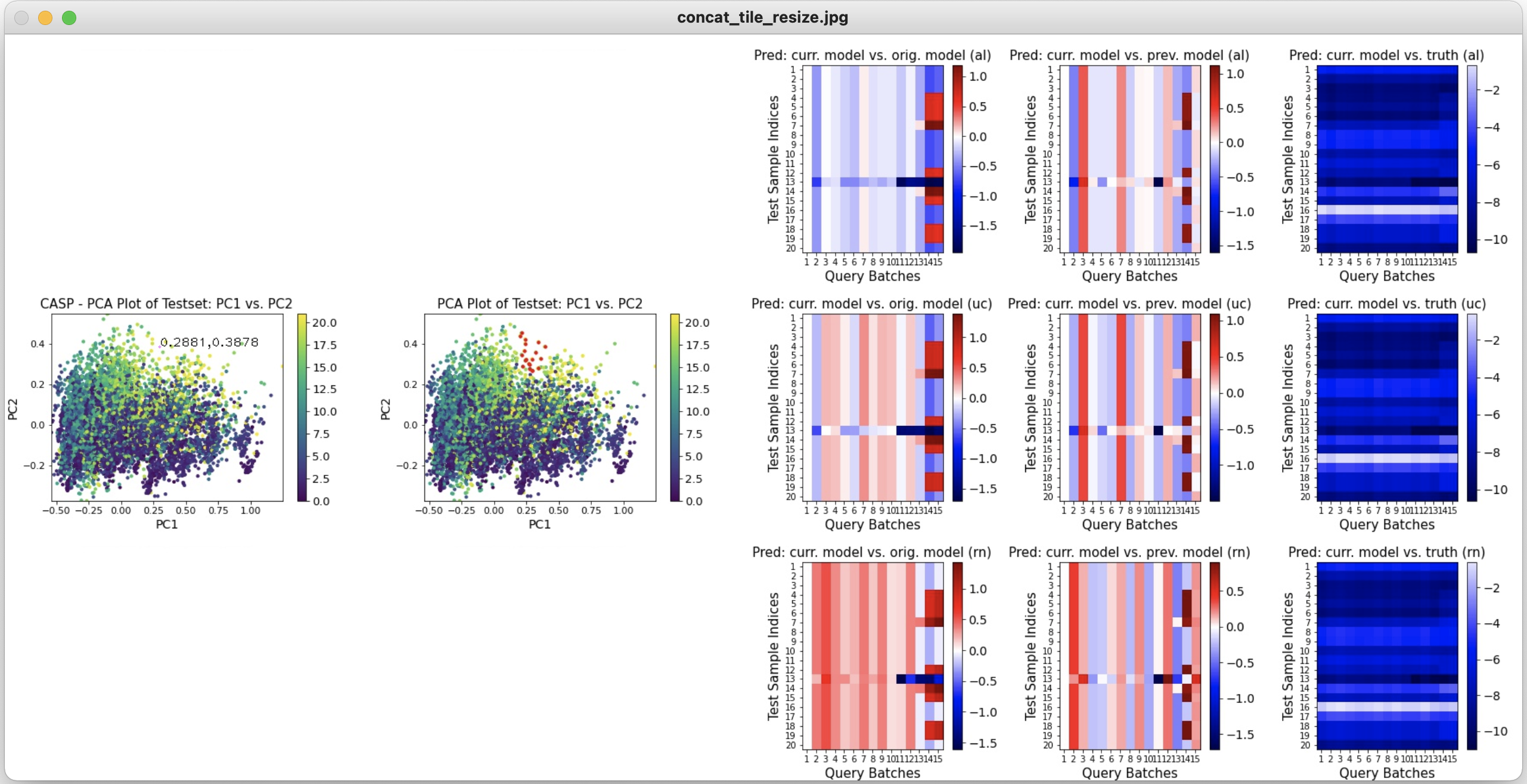}
 {
 }
 \caption{Prediction-change plots for three AL strategies on large value points}
 \centering
\label{fig4}
\end{figure}

\subsection{Distributions of All CASP Data and Queried Samples}

By looking at the truth values of our batch-queried samples, we find that "al" and "uc" tend to query a bit more large value points, compared to the original distribution of all CASP data. However, unfortunately this does not make "al" or "uc" perform much better than "rn" in this large value group.

\begin{figure}[h]

 \includegraphics[width=0.99\linewidth]{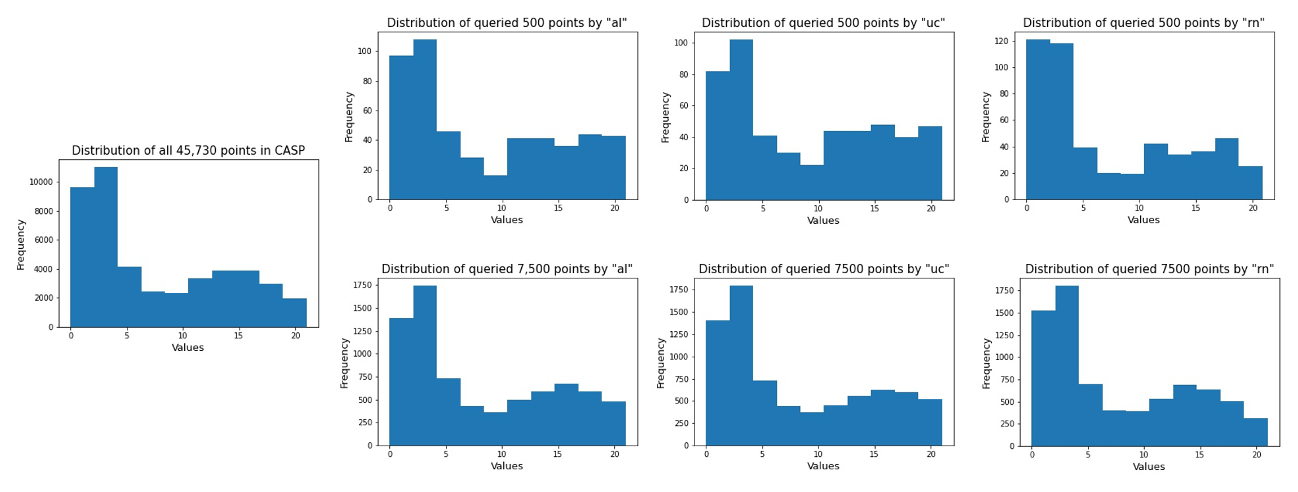}
 {
 }
 \caption{\textbf{Left}: Distribution of y-values for 45,730 points in CASP
          \textbf{Right-Top}: Distribution of 500 queries by three algorithms
          \textbf{Right-Down}: Distribution of 7500 queries by three algorithms
          }
 \centering
\label{fig5}
\end{figure}

\subsection{Demo \& Reproduce All Experiments}

A GitHub demo is available at: \url{https://github.com/anxiousrabbit1/Active_Learning_Visualization_Demo}

To reproduce all experimental results, all data and scripts are available at: \url{https://github.com/anxiousrabbit1/Mondrian_Tree_AL}

\end{document}